\documentclass[conference]{IEEEtran}
\IEEEoverridecommandlockouts

\usepackage{cite}
\usepackage{amsmath,amssymb,amsfonts}
\usepackage{algorithmic}
\usepackage{bm}
\usepackage{booktabs}
\usepackage{graphicx}
\usepackage{multirow}
\usepackage{textcomp}
\usepackage{xcolor}
\usepackage[bookmarks=true]{hyperref}
\usepackage{multicol}
\usepackage{subcaption}
\usepackage{balance}

\def\BibTeX{{\rm B\kern-.05em{\sc i\kern-.025em b}\kern-.08em
    T\kern-.1667em\lower.7ex\hbox{E}\kern-.125emX}}

\let\oldtwocolumn\twocolumn
\renewcommand\twocolumn[1][]{%
    \oldtwocolumn[{#1}{
    \begin{center}
        \includegraphics[width=0.24\linewidth]{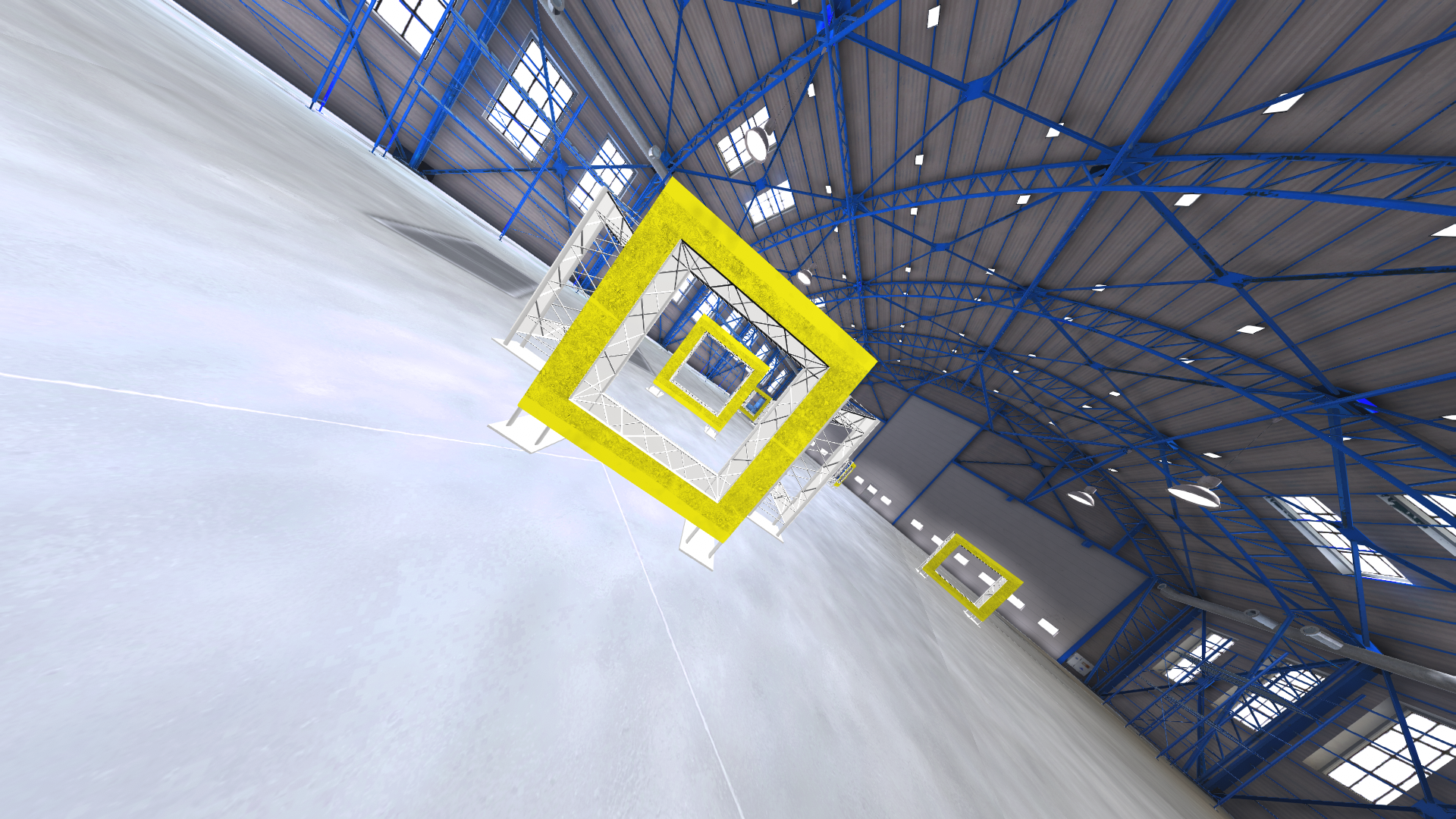}\hfill
        \includegraphics[width=0.24\linewidth]{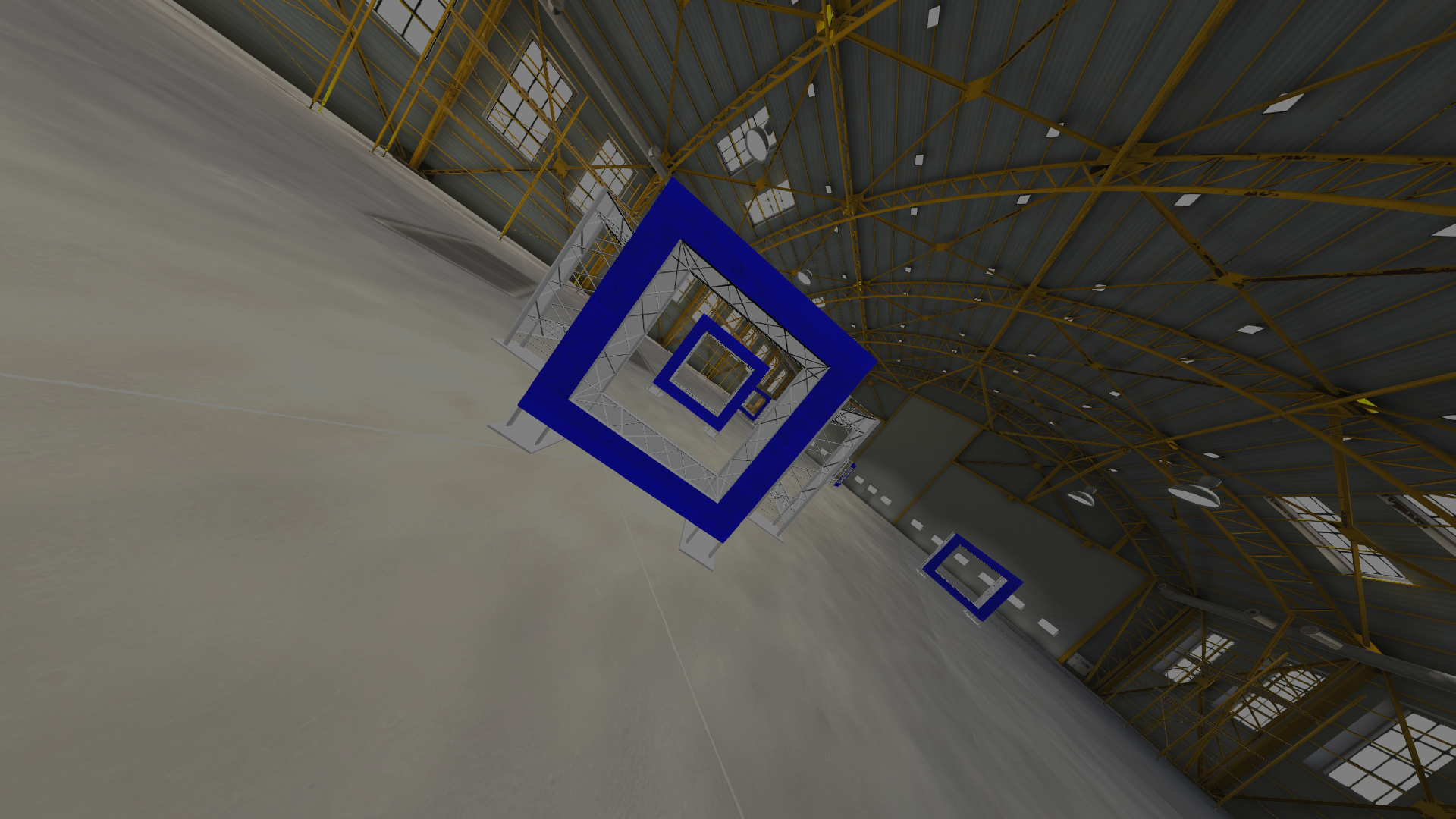}\hfill
        \includegraphics[width=0.24\linewidth]{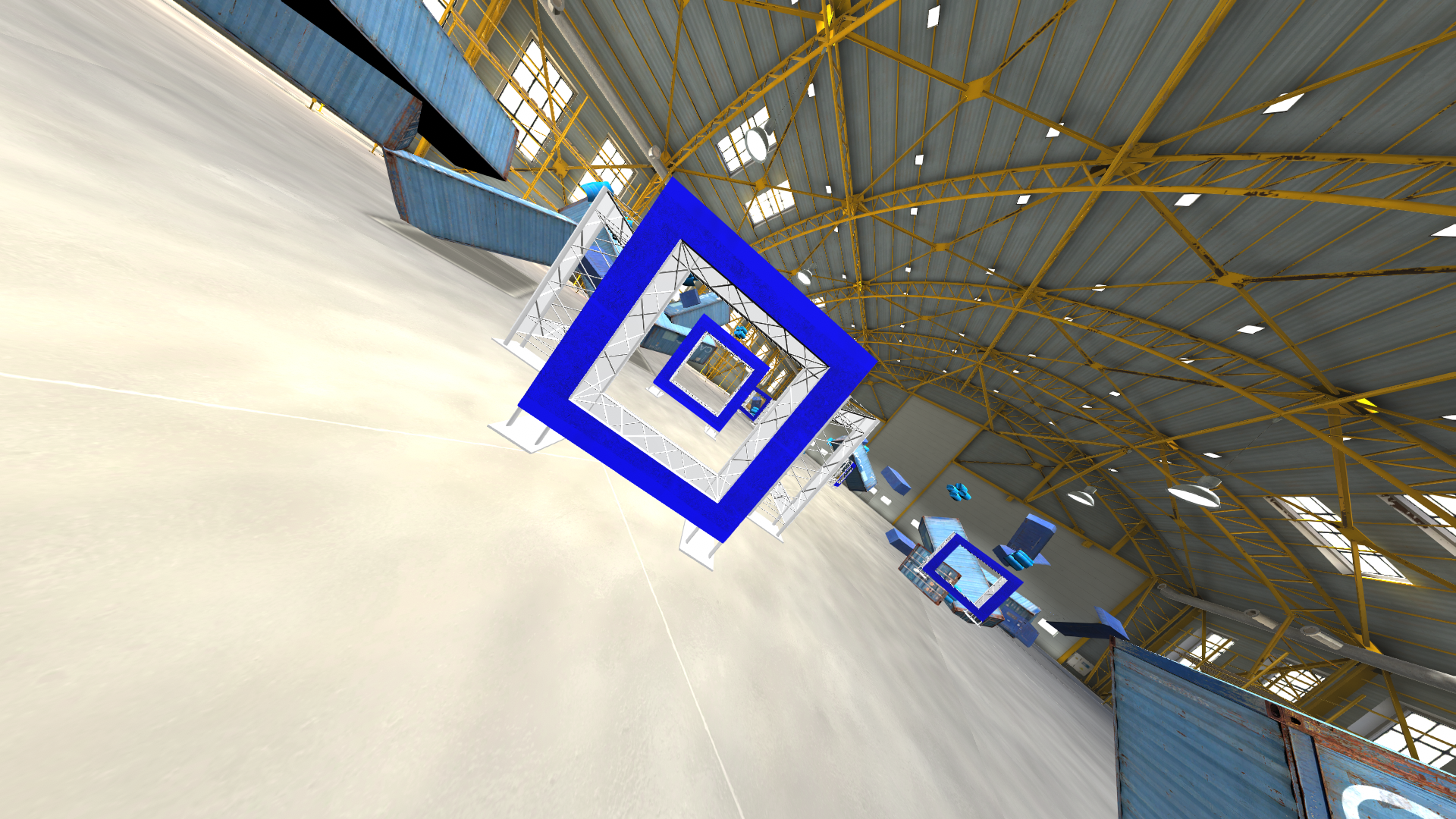}\hfill
        \includegraphics[width=0.24\linewidth]{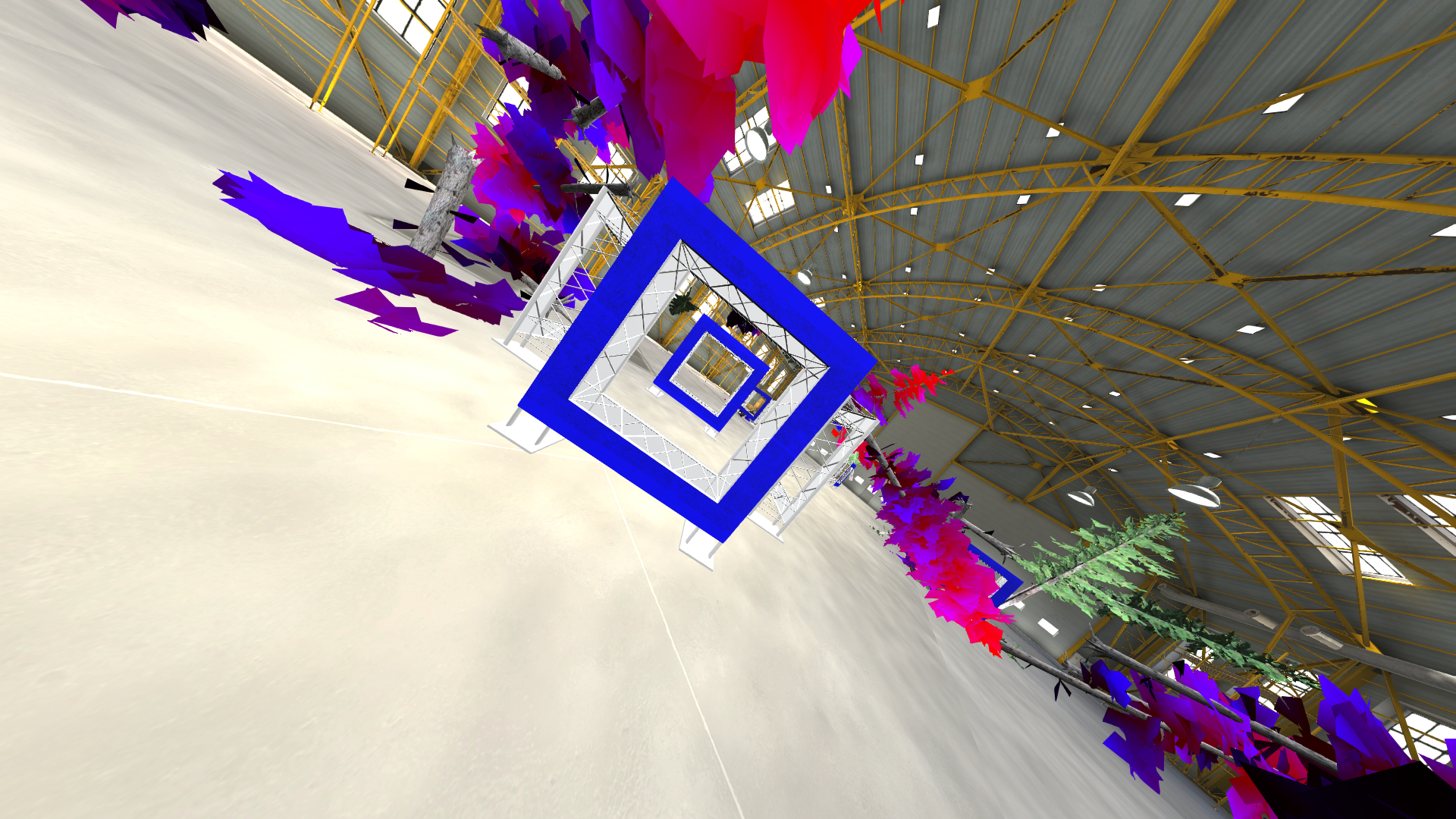}\hfill
        \newline
      \captionof{figure}{We learn deep sensorimotor policies for vision-based autonomous drone racing. The policy maps sensory observations to control commands directly. It achieves results that are robust against various visual disturbances and distractors, which are not experienced during training. From left to right: changing hue value, changing brightness, adding blue boxes, and adding random objects.}
      \label{fig: teaser_img}
    \end{center}
    }
    ]
}

\title{\LARGE \bf
Learning Deep Sensorimotor Policies \\
for Vision-based Autonomous Drone Racing
}

\author{
    Jiawei Fu, Yunlong Song, Yan Wu, Fisher Yu, and Davide Scaramuzza 
    \thanks{
    J. Fu, Y. Song, and D. Scaramuzza are with the Robotics and Perception Group, Department of Informatics, University of Zurich, and Department of Neuroinformatics, University of Zurich and ETH Zurich, Switzerland (\protect\url{http://rpg.ifi.uzh.ch}). 
    Y. Wu and F. Yu are with Visual Intelligence and Systems Group in the Computer Vision Lab at ETH Zurich. 
    This work was supported by the Swiss National Science Foundation (SNSF) through the National Centre of Competence in Research (NCCR) Robotics, the European Union’s Horizon 2020 Research and Innovation Programme under grant agreement No. 871479 (AERIAL-CORE), and the European Research Council (ERC) under grant agreement No. 864042 (AGILEFLIGHT).
    }
}

\begin{document}

\maketitle
\begin{abstract}
Autonomous drones can operate in remote and unstructured environments, enabling various real-world applications. 
However, the lack of effective vision-based algorithms has been a stumbling block to achieving this goal. 
%
Existing systems often require
hand-engineered components for state estimation, planning, and control. 
Such a sequential design involves laborious tuning, human heuristics, and compounding delays and errors. 
This paper tackles the vision-based autonomous-drone-racing problem by learning deep sensorimotor policies. 
We use contrastive learning to extract robust feature representations from the input images and leverage a two-stage learning-by-cheating framework for training a neural network policy.
The resulting policy directly infers control commands with feature representations learned from raw images, forgoing the need for globally-consistent state estimation, trajectory planning, and handcrafted control design.
Our experimental results indicate that our vision-based policy can achieve the same level of racing performance as the state-based policy while being robust against different visual disturbances and distractors.
We believe this work serves as a stepping-stone toward developing intelligent vision-based autonomous systems that control the drone purely from image inputs, like human pilots. 

\vspace{3mm}

Video: \url{https://youtu.be/nPlGR83bC0Q} 

\vspace{1mm}

\end{abstract}


\section{Introduction}
\label{sec: introduction}	

Autonomous drones can travel through complex and dynamic environments at very high speed, holding great potential for a wide range of applications, such as industrial inspection, search and rescue, and reconnaissance. 
Robust vision-based autonomous flight is key to this goal. 
Autonomous vision-based drones have made significant progress in recent years, continuously pushing the vehicle to higher speeds and better robustness.
Several competitions have been organized to push the limit, such as the IROS 2016-19's Autonomous Drone Racing series~\cite{moon2019challenges}, NeurIPS 2019's Game of Drones~\cite{madaan2020airsim}, the 2019 AlphaPilot Challenge~\cite{Foehn20rss,WagterFR2022}, and the ICRA 2022's DodgeDrone Challenge. 

Vision-based autonomous drone racing requires operating the vehicle on the edge of its physical limits, thereby coping with the motion blur and the rapid illumination changes induced by the high speeds and quick rotations of the camera.
The tolerance of the system for mistakes is extremely low: any small error can lead to a crash.

While existing works on vision-based autonomous drone racing rely on globally-consistent state-estimation, planning, and control~\cite{Foehn20rss,moon2019challenges,madaan2020airsim,rojas2021board,WagterFR2022}, human pilots race drones by relying solely on a video stream from the drone's onboard camera, that is, by directly mapping visual input to control commands. While human pilots build a mental model of the drone state, they do not perform any explicit state estimation or trajectory planning~\cite{pfeiffer2021human}. 
In this paper, we make a small step toward emulating human pilots by learning a deep sensorimotor policy for vision-based autonomous drone racing. 

Recent progress in the robot learning community demonstrates that learning deep sensorimotor policies for robotic tasks is feasible.
Methods of this kind usually predict control commands directly from information extracted from high-dimensional sensory inputs.
%
Deep sensorimotor policies have been heavily investigated in many robotic domains, such as object manipulation using robot arms~\cite{levine2016end, akkaya2019solving} or benchmark control of simulated robots~\cite{tassa2018deepmind, laskin2020reinforcement, hansen2021generalization}.
This line of works has the advantage that the policy algorithm relaxes the need for a globally-consistent state information and enlarges the application of the system.
However, learning deep sensorimotor policies for vision-based navigation still faces several challenges, including high sample complexity and poor generalization.

%
%
%
%
%
%

%

\newpage

An overview of our system is given in Fig.~\ref{fig: overview}. 
Our main contribution is a deep sensorimotor policy that can jointly solve perception, planning, and control for autonomous drone racing, without relying on an globally-consistent state of the drone nor on trajectory planning. The inputs to our policy are a sequence of images and part of the drone state (orientation, velocity, acceleration) but \emph{no globally-consistent position} information.

Our method consists of two key components: privileged policy training and robust feature learning. 
First, we leverage a two-stage learning-by-cheating framework for policy training. 
Second, we use contrastive learning and data augmentation to extract robust image embeddings from RGB images.
%

%
Furthermore, we compare our vision-based deep sensorimotor policy against a neural control policy that utilizes the full globally-consistent state information~\cite{song2021autonomous}.
Our experiments, conducted in a realistic simulator~\cite{yunlong2020flightmare}, show that our vision-based deep sensorimotor policy achieves the same level of racing performance while being robust against different visual disturbances and distractors. 
Finally, we benchmark the performance of our vision-based policy against the time-optimal trajectory generation algorithm~\cite{Foehn2021science}, which offers a theoretical minimum time. 
Our policy achieves lap times close to the time-optimal solution.

%
\section{Related Work}
\label{sec: relatedwork}	


%
Different approaches have been studied to tackle autonomous drone racing. 
State-based methods that rely on globally accurate position information have been used extensively. 
Foehn et al.~\cite{Foehn2021science} presented the time-optimal trajectory generation by jointly improving the time allocation and the trajectory. The algorithm enabled them to outperform human experts in drone racing. 
In \cite{hwangbo2017control,song2021autonomous, nagami2021hjb}, authors used reinforcement learning to train a neural network as the policy.
For example, Song et al.~\cite{song2021autonomous} utilized relative gate positions towards the next gates to achieve near-time-optimal performance. 
Nagami et al.~\cite{nagami2021hjb} initialized a network by mimicking a simplified controller and further trained it with reinforcement learning. The hierarchy allowed the policy to outperform a trajectory planning policy.
Although promising results can be generated by state-based methods, the assumption of exact position information limits the application of the methods.

Prior work on vision-based drone racing decouples the perception, planning, and control modules. 
In the work of Foehn et al.~\cite{Foehn20rss}, visual-inertial odometry~(VIO) was fused with a CNN-based gate corner detection for robust state estimation. 
A receding horizon path planner generates a time-optimal trajectory using motion primitives based on a point-mass model of the drone platform.
However, the point-mass assumption cannot represent the true actuation limits of the drone and may lead to dynamically infeasible trajectories. 
In~\cite{kaufmann2018deep, Loquercio19TRO, wang2021robust}, authors first use data-driven methods to train the neural networks that can predict the waypoint and the desired speed.
%
Afterward, a minimum jerk trajectory is planned for passing through the waypoint and then tracked by a low-level controller. 
%
%
Muller et al.~\cite{muller2019learning} propose to train a neural network for local trajectory planning, in which a downstream control policy is used to track the trajectory and generate low-level commands for vehicle control. 
The trajectory labeling requires additional engineering efforts and can result in ambiguity as each image can be labeled with different trajectories.
The decoupling of the perception, planning, and control modules inevitably involves simplified assumptions or manual design of parameters, leading to sub-optimality during high-speed flight.

Recent advance in data-driven control~\cite{kaufmann2020RSS, hwangbo2019learning, LeeSR2020, zhang2021end} indicates the potential of developing autonomous systems using sensorimotor control, in which a neural network policy can map high-dimensional sensory inputs directly to control commands. 
However, with naive training, the deep sensorimotor policy might suffer from poor generalization when facing unseen disturbances.
Different approaches have been employed to alleviate the overfitting issue, such as data augmentation~\cite{hansen2021generalization, james2019sim}, injecting known biases~\cite{wang2021unsupervised}, and extracting invariant information~\cite{li2021domain}.
Most of them are applied to video games~\cite{wang2021unsupervised, laskin2020reinforcement}, robot arm control~\cite{james2019sim, lee2019network}, or autonomous driving~\cite{fan2021secant}. 
The generalization capability of a neural network policy for minimum-time flight has drawn much less attention due to several challenges, such as low reaction time and a rapid change of the image observations. 

\section{Methodology}
\label{sec: method}	

\begin{figure*}[htbp]
  \centerline{\includegraphics[width=1\textwidth]{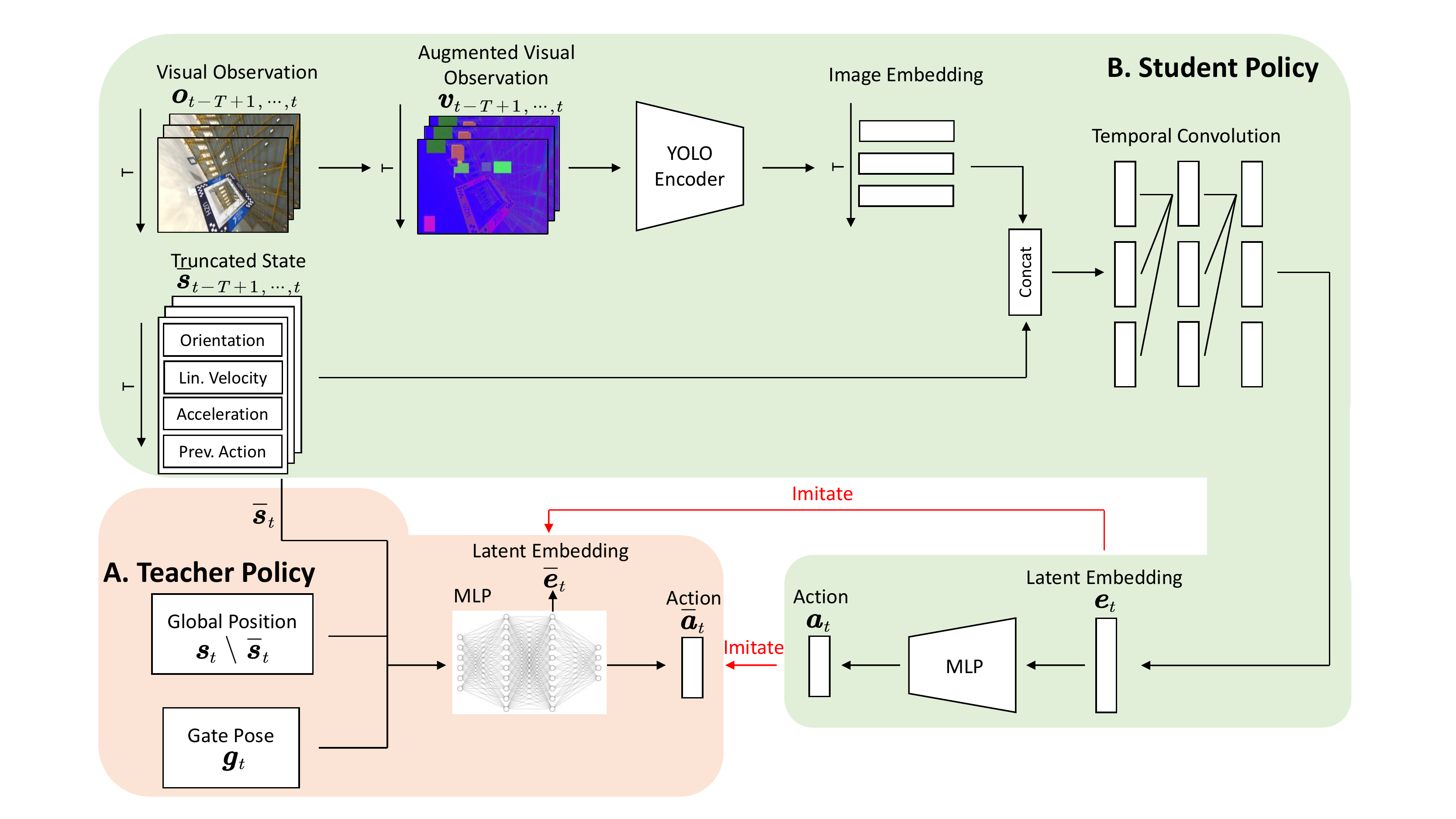}}
  \caption{Overview of our policy training method. We first train a teacher policy with access to privileged state information using model-free reinforcement learning. This teacher policy is then distilled into a student policy, which is trained to do perception, planning, and control jointly.}
  \label{fig: overview}
\end{figure*}

An overview of our method is visualized in Fig.~\ref{fig: overview}. 
Our approach consists of two key components: policy training and feature learning. 
The policy training is done using privileged reinforcement learning and imitation learning, where a student policy mimics the action of a teacher policy. 
To process high-dimensional image data and allow efficient policy training, we use YOLO~\cite{redmon2016you, glenn_jocher_2022_6222936} to extract low-dimensional image embeddings. 

\subsection{Policy Training}
\label{sec:policy-training}

\textbf{Teacher Policy Training:}
The first step is to obtain a state-based teacher policy that can push the vehicle to its maximum performance. 
We use reinforcement learning to train a multilayer perceptron (MLP) policy~$\pi_\text{teacher}$ for passing through a sequence of gates $\{G_i |, i=1, \cdots, N\}$ in minimum time. 
At every time step $t$, the agent is at state $\bm{s}_t$ and receives information about the gate state $\bm{g}_t$. 
Our goal is to find the optimal policy $\pi^{\ast}_\text{teacher}$ that maximizes the expected discounted return: $\pi^{\ast}_\text{teacher} = \operatorname*{arg\,max}_\pi \mathbb{E} \left[ \sum_{t=0}^{T-1}  \gamma^t r_t \right]$, where $\gamma$ is the discount factor and $r_t$ is the reward at time stage $t$. 
In privileged learning, the teacher policy has access to all ground truth information, including the vehicle's state $\bm{s}_t$ and the gate state $\bm{g}_t$. 
Hence, the teacher policy generates an action $\bar{\bm{a}}_t \sim \pi(\bm{s}_t, \bm{g}_t)$ given both states. 
The policy outputs control commands in the form of mass-normalized collective thrust and angular velocity. 

The main objective is to minimize the lap time, which is equivalent to maximizing the path progress along the center line connection between two consecutive gates~\cite{song2021autonomous}. 
In addition, we maximize a perception-aware reward to maximize the visibility of the next gate. 
The perception-aware reward incentivizes the policy to face the camera toward the next passing gate, which is crucial for vision-based flight since our environment is only partially observable when using a camera. We denote the position and velocity of the center of the next gate on the image plane by $\bm{p}_c$ and $\bm{\dot{p}}_c$, respectively. The perception award reward is formulated to keep the gate in the image center and reduce the motion blur~\cite{falanga2018pampc} as
$
r_p=\exp\left( -||\bm{p}_c||_2 - ||\bm{\dot{p}}_c||_2 \right),
$
where $||\cdot||_2$ refers to 2-norm.
\\

\textbf{Student Policy Training:}
After we obtain a teacher policy $\pi^{\ast}_\text{teacher}(\bm{s}_t, \bm{g}_t)$ that can race the drone optimally, we distill the teacher's knowledge to a student policy $\pi_\text{student}(\bm{\overline{s}}_t, \bm{o}_t)$ that does not have access to the privileged information about the environment. 
Specifically, the student policy can only observe part of the drone state $\bm{\overline{s}}_t$, which does not contain the vehicle's global position, and need to infer the gate information from the camera observation $\bm{o}_t \rightarrow \bm{g}_t$. 
There are three key components of our student policy: a feature extractor, a memory-based neural network, and a policy network. 

We use YOLO~\cite{redmon2016you, glenn_jocher_2022_6222936} as the feature extractor and train it to detect all gates in a given image. 
We use average pooling to downsample the output of the three convolutional layers in its detection head and concatenate them as the embedding of the image $\bm{z}_t$.
Since the detection head is the rightmost module of YOLO, this embedding contains all the information for detecting the gates, and hence, is a sufficient for representing the image. 
Note that we additionally normalize the embedding with $l$2-normalization, which is empirically found to be beneficial for the convergence of policy training.
%


When using a single camera, the environment becomes a partially observable environment.
To this end, we use a temporal convolutional network (TCN)~\cite{bai2018empirical} for the policy representation. 
The embedding from the image is concatenated with the truncated vehicle state $\bm{\overline{s}}_t$. 
The sequence of concatenated embeddings is then fed into the TCN to extract temporal information from history observations.
Finally, we use a MLP to regress the control command. 
The MLP takes the output of the TCN as input and produces the student policy's action which is of the same format as the teacher policy. 

We use imitation learning to train the student policy. We define an action loss $\mathcal{L}_A$ that is the mean square error between the outputs of the teacher policy and the student policy.
\begin{equation}
\mathcal{L}_A(\bm{\theta}) =||\pi_\text{student}(\bm{\overline{s}}_t, \bm{o}_t | \bm{\theta}) -\pi^{\ast}_\text{teacher}(\bm{s}_t, \bm{g}_t)||_2.
\end{equation}
In addition, to better enable the knowledge transfer between the teacher policy and the student policy, we also add a latent loss $\mathcal{L}_E$ to supervise the output of the TCN $\bm{e}_t$ with the intermediate embedding of the teacher $ \bm{\bar{e}}_t$, written as $
\mathcal{L}_E(\bm{\theta})=||\bm{\bar{e}}_t - \bm{e}_t||_2
$.
Therefore, we minimize the total loss for the imitation learning
\begin{equation}
\min_{\bm{\theta}} \quad \mathcal{L}=\mathcal{L}_A (\bm{\theta})+\lambda \mathcal{L}_E(\bm{\theta}),
\end{equation}
where $\lambda$ is a coefficient to weight the latent loss.

\subsection{Robust Feature Learning via Data Augmentation}
\label{sec: method-feature-learning}

%

\begin{figure}[htbp]
  \centerline{\includegraphics[width=0.4\textwidth]{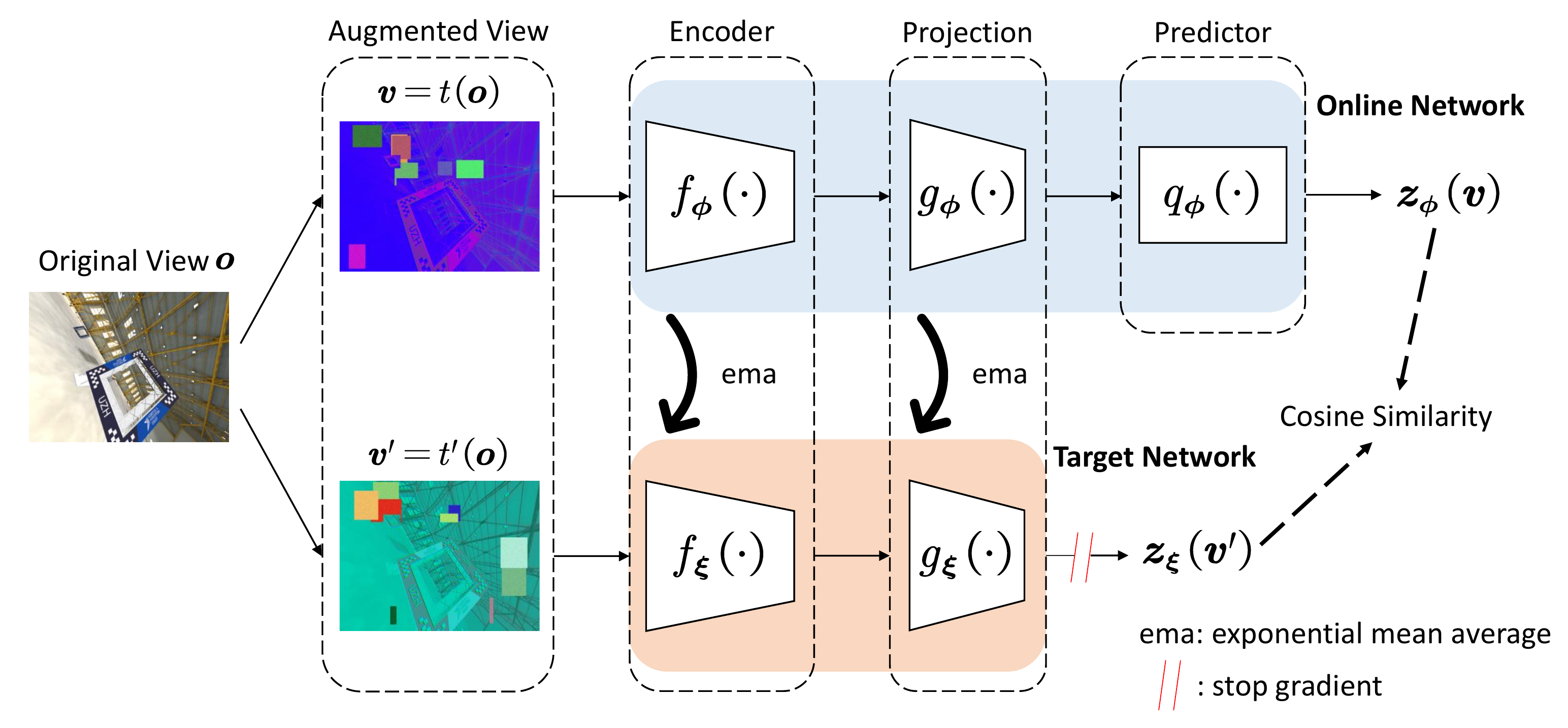}}
  \caption{Contrastive learning framework~\cite{grill2020bootstrap}. }
  \label{fig: byol}
\end{figure}

To learn robust image embeddings against disturbance, we train the encoder with contrastive learning. We use the framework introduced in ~\cite{grill2020bootstrap}. 
The framework (shown in Fig.~\ref{fig: byol}) contains an online network and a target network. 
The online network defined by parameters~$\bm{\phi}$ contains three components: an encoder $f_{\bm{\phi}}$, a projection $g_{\bm{\phi}}$, and a predictor $q_{\bm{\phi}}$. 
The target network is an exponential mean average of the online network and defined by parameters $\bm{\xi}$. It is comprised of two components: a target encoder $f_{\bm{\xi}}$ and a target projection $g_{\bm{\xi}}$. 

An input image $\bm{o}$ is passed through two augmentations, denoted as $t$ and $t'$, to obtain two augmented views $\bm{v}=t(\bm{o})$ and $\bm{v}'=t'(\bm{o})$ separately. 
Then the embedding prediction $\bm{z}_{\bm{\phi}}(\bm{v})=q_{\bm{\phi}}(g_{\bm{\phi}}(f_{\bm{\phi}}(\bm{v})))$ is extracted by the online network while the embedding target $\bm{z}_{\bm{\xi}}(\bm{v}')=g_{\bm{\xi}}(f_{\bm{\xi}}(\bm{v}'))$ is extracted by the target network. 
A cosine similarity loss is applied to align the embeddings,
\begin{equation}
\mathcal{L}^{cos}(\bm{v},\bm{v}')=2-2\cdot \frac{<\bm{z}_{\bm{\phi}}(\bm{v}),\bm{z}_{\bm{\xi}}(\bm{v}')>}{||\bm{z}_{\bm{\phi}}(\bm{v})||_2 \cdot ||\bm{z}_{\bm{\xi}}(\bm{v}')||_2}
\end{equation}
where $||\cdot||_2$ denotes $l$2-normalization. 
%
%
%

%

\section{Experiments}
\label{sec: experiments}

We design our experiments to answer the following research questions: 
1) our student policy does not depend on the vehicle's global position and can only observe the environment partially; how does such a vision-based control system compare to a state-based system?
2) our policy is trained with data augmentation. Can the data augmentation align image embeddings given 
different visual disturbances and distractors, and is the policy robust against those disturbances?
3) our policy still relies on some part of the vehicle state; how well can the policy handle estimation errors in the drone state?

\subsection{Experimental Setup}
\begin{figure*}[!htp]
\centering
\setlength{\tabcolsep}{0.5em}
\begin{tabular}{ccc}
 \includegraphics[width=0.3\linewidth]{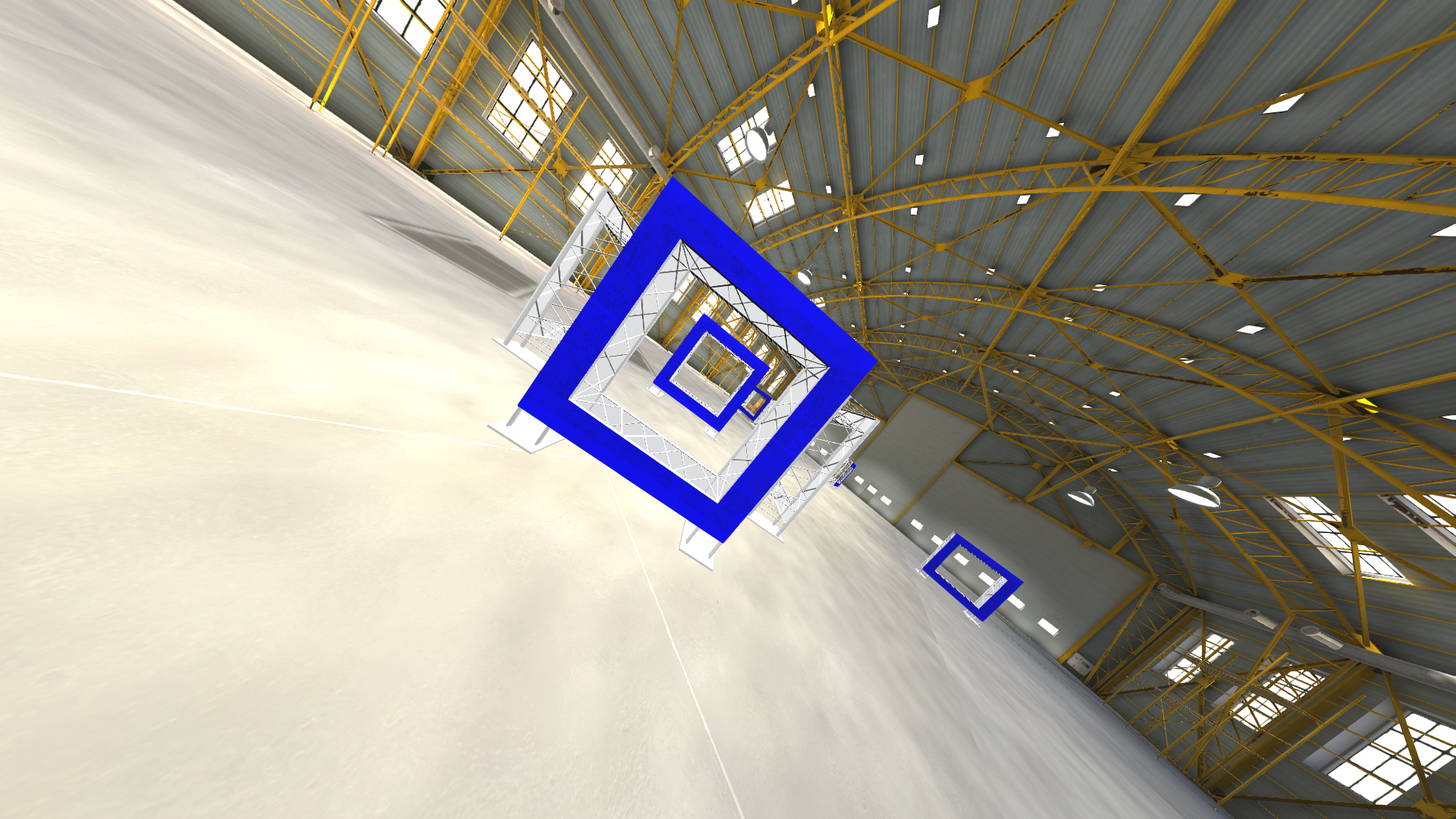}
 &
 \includegraphics[width=0.3\linewidth]{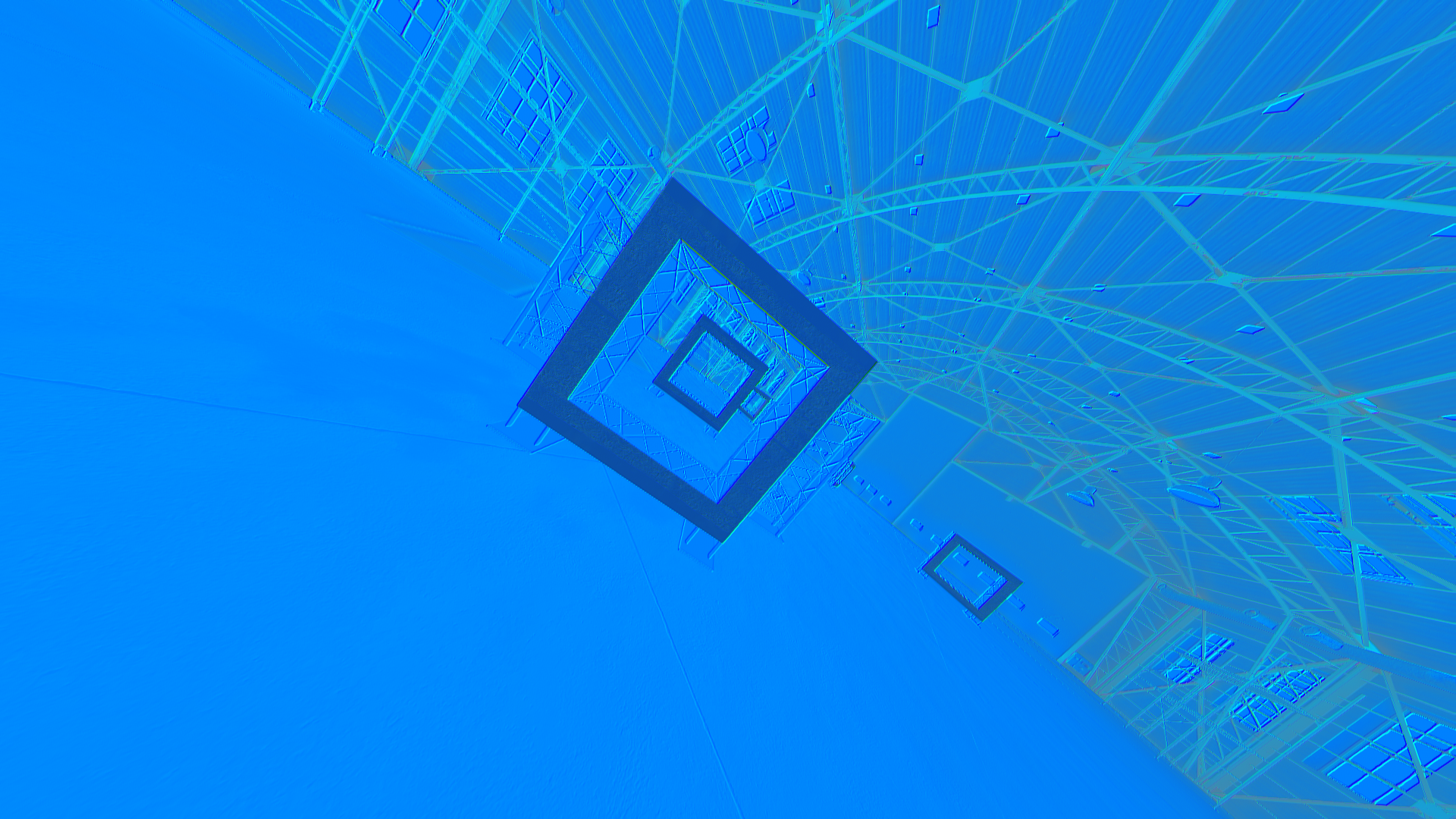}
 &
 \includegraphics[width=0.3\linewidth]{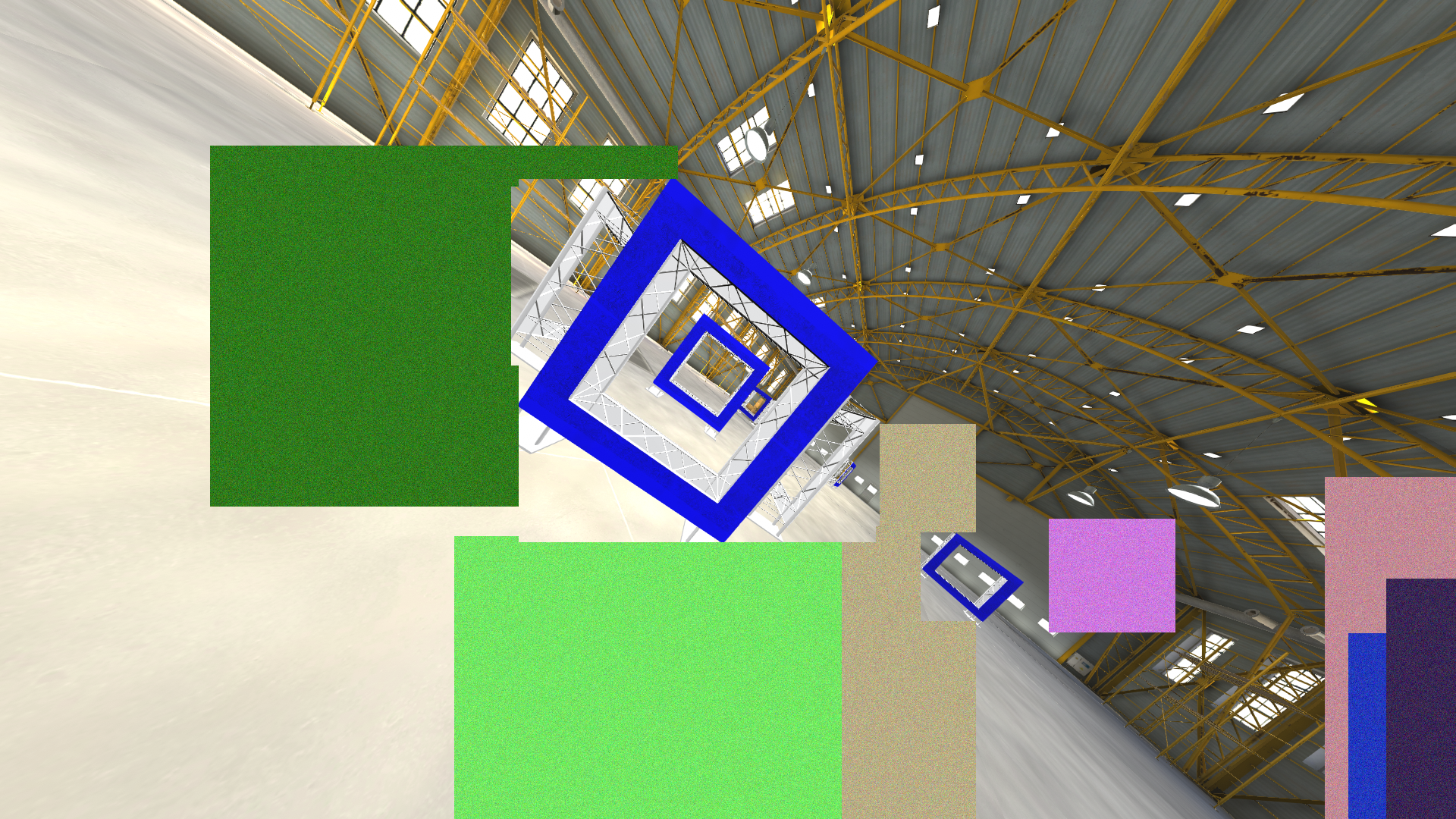} \\
\end{tabular}
\caption{Visualization of data augmentations used during training. Left: no augmentation. Middle: random convolution. Right: random cutout-color.}
\label{fig: large_aug}
\vspace{-2mm}
\end{figure*}

\textbf{Simulator Environment:} We conduct experiments using the Flightmare~\cite{yunlong2020flightmare} simulator, a realistic quadrotor simulator with various racing tracks and realistic racing environments.
We set up three different race tracks (Circle, Figure8, and SplitS) in a warehouse environment (see the visualization in Fig.~\ref{fig: large_aug} left). 
For training the teacher policy, we use a customized implementation of the proximal policy optimization algorithm~(PPO)~\cite{schulman2017proximal} based on the code from~\cite{stable-baselines3}.  
For training the student policy, we implement an imitation learning pipeline. 
%
For learning robust feature representations from raw images, we use data augmentation with random convolution and random cutout-color (see Fig.~\ref{fig: large_aug} middle and right).

\textbf{Evaluation:} To evaluate our policy, we rollout for 10 episodes, with quadrotor starting from different starting positions, which are sampled from a uniform distribution between -0.1m and 0.1m in $x$, $y$, $z$-axis of each. 
We evaluate the performance of our policy using two different metrics: Lap Time and Success Rate.
The lap time indicates the racing performance of our policy, while the success rate indicates the robustness of the policy. 
We report the lap time by computing the time required by the policy to finish one complete track and calculate the success rate by calculating the ratio that the quadrotor can successfully finish one full lap without crashing among the ten rollouts.

\subsection{Baseline Comparisons}
\begin{figure}[!htp]
\centering
\setlength{\tabcolsep}{0.5em}
\begin{tabular}{ccc}
 \includegraphics[width=0.3\linewidth]{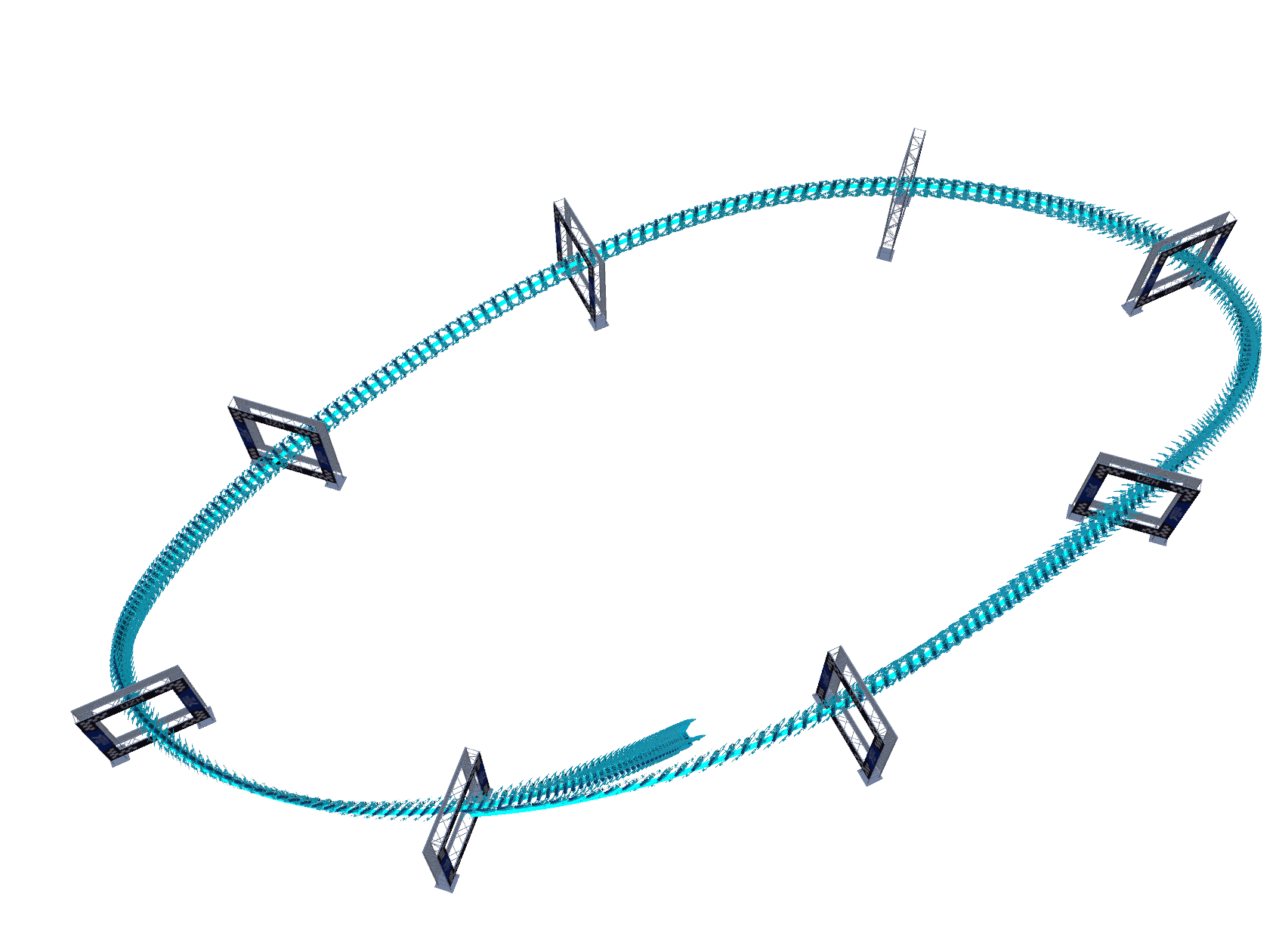}
 &
 \includegraphics[width=0.3\linewidth]{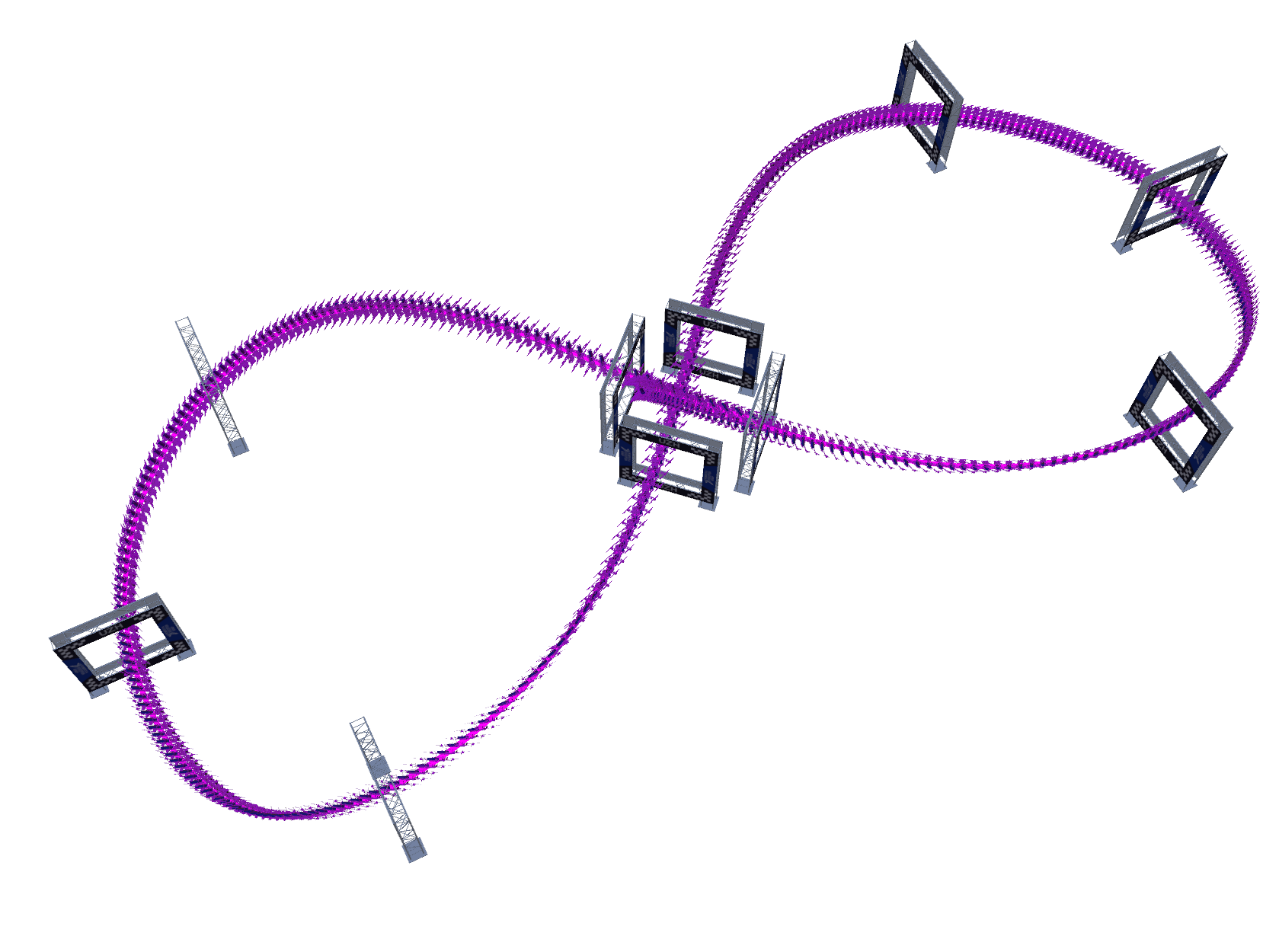}
 &
 \includegraphics[width=0.3\linewidth]{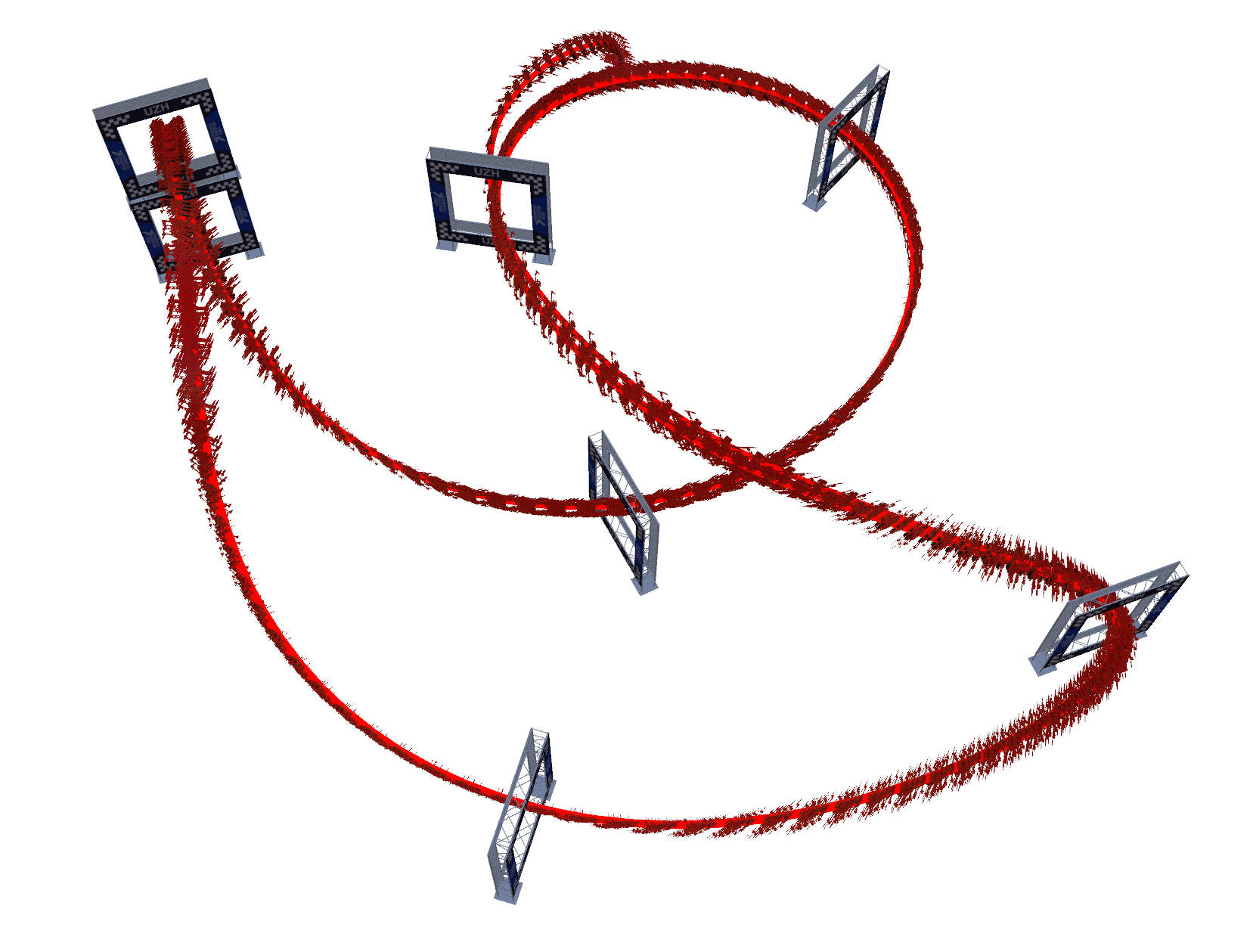} \\
\end{tabular}
\caption{Visualization of trajectories. Left: Circle. Middle: Figure8. Right: SplitS.}
\label{fig: traj_vis}
\vspace{-2mm}
\end{figure}

We compare our vision-based policies against two baselines: a state-based learning-based policy~\cite{song2021autonomous} and a time-optimal trajectory~\cite{Foehn2021science}. 
%
The state-based policy controls the drone using ground truth information about the drone state, including position, velocity, orientation, and acceleration, as well as the pose of the next two gates. 
The time-optimal trajectory serves as the theoretical minimum bound for our platform. 
%
The student policy does not have access to ground truth information about the drone's position and the gate poses. 
Instead, it uses a camera to capture RGB images and controls the drone directly using the image. 
Therefore, the student policy can only observe the environment partially, similar to how human pilots control the drone using the first-person-view camera. 
The result is shown in Table~\ref{tab:baseline}. 
A visualization of the trajectories is given in Fig.~\ref{fig: traj_vis}.
Both policies achieve strong performance on three different race tracks with high success rates. 
The student policy learns to cut corners, resulting in lower lap time, but more risky behaviors. 

\begin{figure}[htbp]
  \centerline{\includegraphics[width=0.35\textwidth]{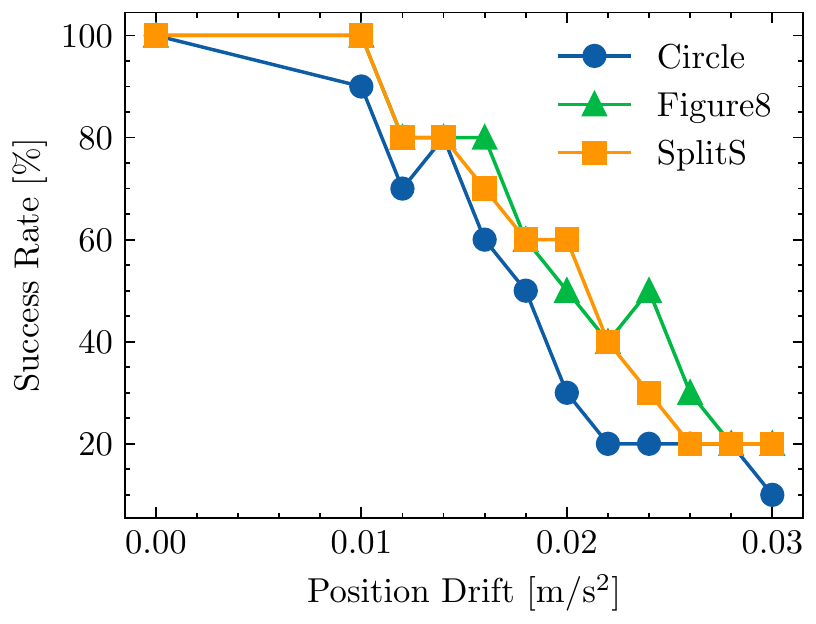}}
  \caption{Success rates of the state-based policy over position drift. }
  \label{fig: drift}
\end{figure}


In reality, the vehicle states are prone to error due to the drift in state estimation and measurement errors.  
Despite impressive results in visual-(inertial) odometry in recent years, high-speed flight with six degrees of freedom motion remains challenging for existing
estimation algorithms~\cite{delmerico19icra}. 
Hence, the state-based control system is subject to failure since the policy relies heavily on position estimation.
We investigate this problem using a simulated VIO pipeline, in which we simulate position drift. 
Fig.~\ref{fig: drift} shows how position drifts affect the performance of the state-based policy. 
Given perfect state information, the policy achieves 100~$\%$ success rates on all tracks. 
However, as we increase the drift in position, the success rates collapse quickly. The VIO drift can be alleviated by relocalizing with respect to the gates but this is challenging because the camera suffers from motion blur and limited field of view.
On the other hand, our vision-based policy is not affected by the drift in the position since it does not rely on that information. 

\begin{table*}
    \centering
    \begin{tabular}{cccccccc}
    \toprule
         & \multicolumn{3}{c}{Lap Time [s]} && \multicolumn{3}{c}{Success Rate} \\
         \cline{2-4} \cline{6-8}
         & Circle & Figure8 & SplitS && Circle & Figure8 & SplitS \\
         \midrule
         Time-optimal Trajectory~\cite{Foehn2021science} & 4.68 & 6.26 & 7.93 && - & - & - \\
         State-based Policy~\cite{song2021autonomous} & 4.97$\pm$0.01 & 6.84$\pm$0.05& 8.74$\pm$0.01 && 1.0 & 1.0 & 1.0 \\
         Vision-based Policy~(ours) & 4.95$\pm$0.01 & 6.76$\pm$0.01& 8.58$\pm$0.01 && 1.0 & 1.0 & 1.0 \\
         \bottomrule
    \end{tabular}
    \caption{Performance of the time-optimal trajectory, state-based teacher policy, and vision-based student policy on three different race tracks.}
    \label{tab:baseline}
\end{table*}

\subsection{Handling Visual Disturbances and Unseen Distractors}


We deploy our vision-based system in various unseen contexts to investigate how it performs against unseen visual disturbances, such as environments with color changes, brightness changes, and environments with many randomly arranged unseen objects.
We darken the environment by lowing the brightness value from 1 to 0.5 and 0.8, and we also change environment colors by tuning the image hue value from 0 to both 0.5 and -0.5. Fig.~\ref{fig: teaser_img} left and middle-left provide examples of environment with brightness values of 0.5 and hue values of 0.5, respectively. In addition, we also place some visual distractors randomly around the environment, including blue boxes that are similar in color and shape to the racing gates (see visualization in Fig.~\ref{fig: teaser_img} middle-right), and some random objects with irregular shapes (see visualization in Fig.~\ref{fig: teaser_img} right).

As presented in Table~\ref{tab:color_brightness_change}, our system is robust against various types of visual disturbances while still maintaining a high success rate and comparable lap time on all three racing tracks, which demonstrates the effectiveness of our image feature learning mechanism. 
In the following section, we further investigate how the image encoder trained using contrastive learning can generalize to these visual disturbances.

\begin{table*}
\centering
\begin{tabular}{ccccclccc}
\toprule
\multicolumn{2}{c}{\multirow{2}{*}{}} & \multicolumn{3}{c}{Lap Time {[}s{]}}          &  & \multicolumn{3}{c}{Success Rate} \\ \cline{3-5} \cline{7-9} 
\multicolumn{2}{c}{}                  & Circle        & Figure8       & SplitS        &  & Circle    & Figure8   & SplitS   \\ \midrule
\multirow{2}{*}{\textbf{Brightness Change}}   & 0.5   & 4.88$\pm$0.01 & 6.71$\pm$0.01 & 8.60$\pm$0.01 &  & 1.0       & 0.8       & 1.0      \\
                              & 0.8   & 4.91$\pm$0.02 & 6.68$\pm$0.01 & 8.65$\pm$0.01 &  & 1.0       & 1.0       & 1.0      \\ \midrule
\multirow{2}{*}{\textbf{Hue Change}} & -0.5  & 4.91$\pm$0.01 & 6.72$\pm$0.03 & 8.66$\pm$0.01 &  & 1.0       & 1.0       & 1.0      \\
                              & 0.5   & 4.92$\pm$0.02 & 6.71$\pm$0.01 & 8.65$\pm$0.01 &  & 1.0       & 1.0       & 1.0      \\ \midrule
\multirow{2}{*}{\textbf{Blue Boxes}} & 10 & 4.94$\pm$0.01 & 6.87$\pm$0.01 & 8.64$\pm$0.01 &  & 1.0       & 1.0       & 1.0      \\
& 60 & 4.90$\pm$0.02 & 6.77$\pm$0.01 & 8.73$\pm$0.01 &  & 1.0       & 1.0       & 0.9      \\ \midrule
\multirow{2}{*}{\textbf{Random Objects}} & 10  & 4.94$\pm$0.02 & 6.75$\pm$0.02 & 8.67$\pm$0.01 &  & 1.0       & 1.0       & 1.0      \\
                              & 60   & 4.91$\pm$0.02 & 6.81$\pm$0.03 & 8.67$\pm$0.01 &  & 1.0       & 0.6       & 1.0      \\ \bottomrule
\end{tabular}
\caption{Success rate and lap time of the vision-based sensorimotor control policy. Our policy is robust against different visual disturbances, including brightness change (darker environment with brightness value changing from 1 to 0.5 and 0.8, respectively) and hue change (environment color changes with hue value tuning from 0 to 0.5 and -0.5, respectively), and the presence of various distractors, including blue boxes and random objects, of various densities (10 and 60 distractors are added, respectively).}
\label{tab:color_brightness_change}
\end{table*}



\subsection{Aligning Image Embeddings}
\begin{figure}[!htp]
\centering
\setlength{\tabcolsep}{0em}
\begin{tabular}{l ccc}
 &  Circle Track & Figure8 Track & Split-S Track \\
 \hline
A 
 & \includegraphics[width=0.25\linewidth]{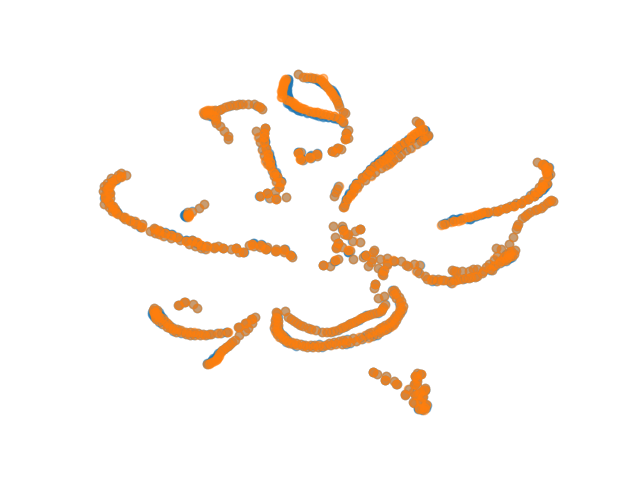} &
 \includegraphics[width=0.25\linewidth]{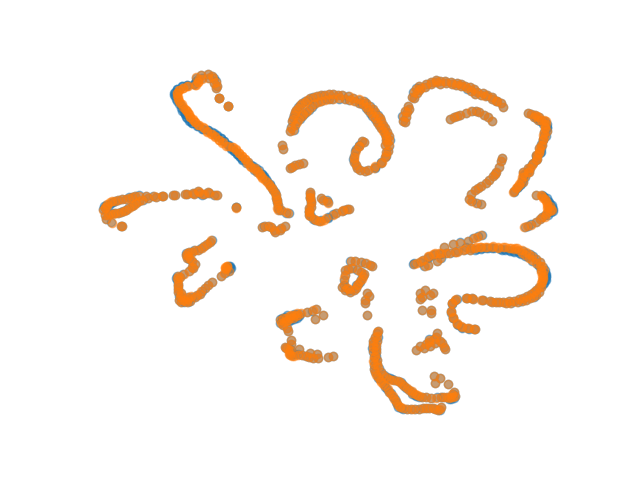} &
 \includegraphics[width=0.25\linewidth]{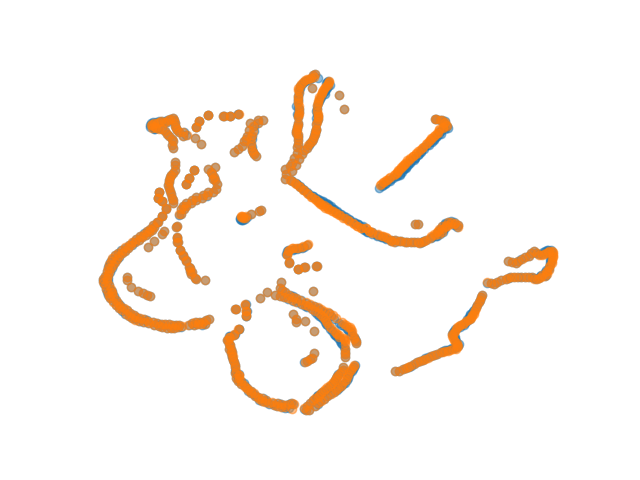} \\
 \hline
B  
& \includegraphics[width=0.25\linewidth]{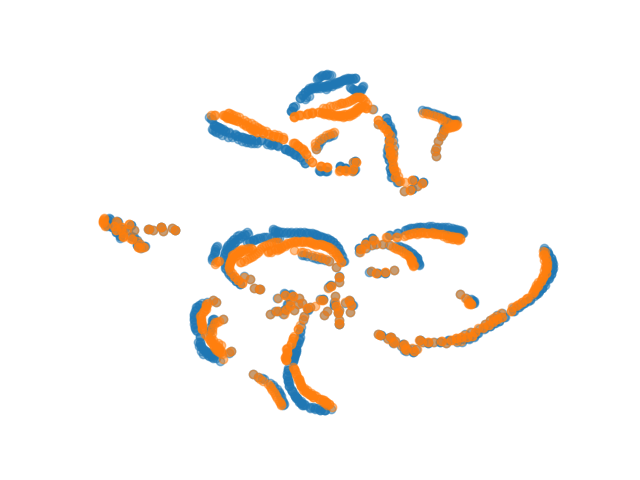} &
 \includegraphics[width=0.25\linewidth]{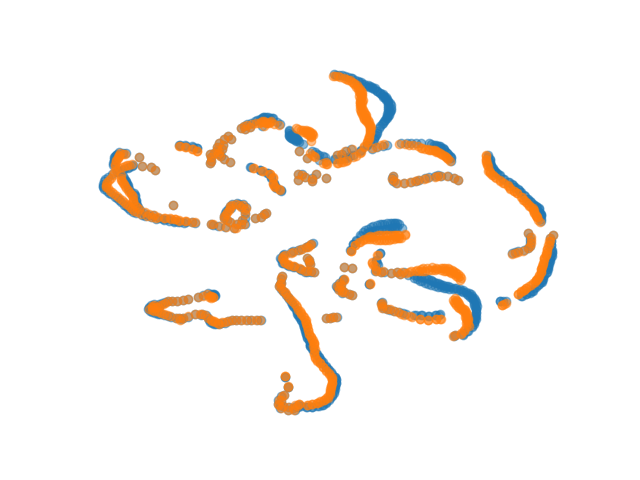} &
 \includegraphics[width=0.25\linewidth]{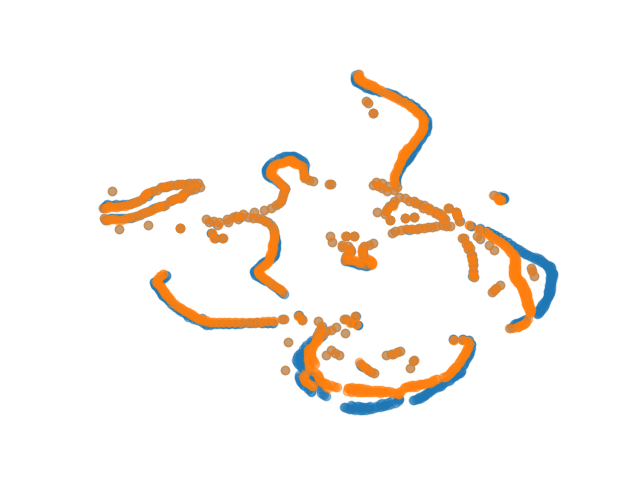} \\
 \hline
C
 & \includegraphics[width=0.25\linewidth]{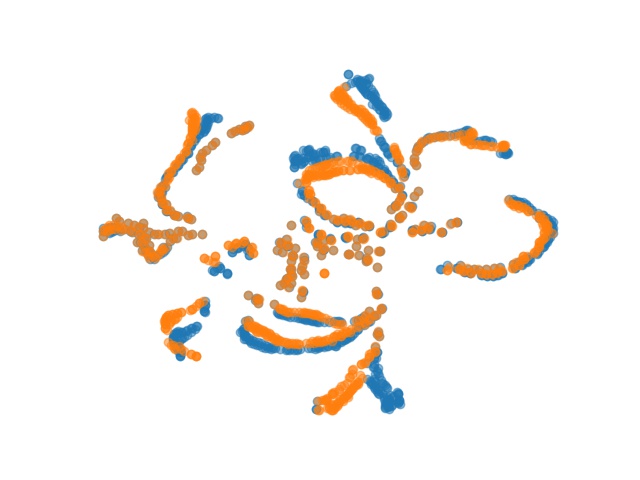} &
 \includegraphics[width=0.25\linewidth]{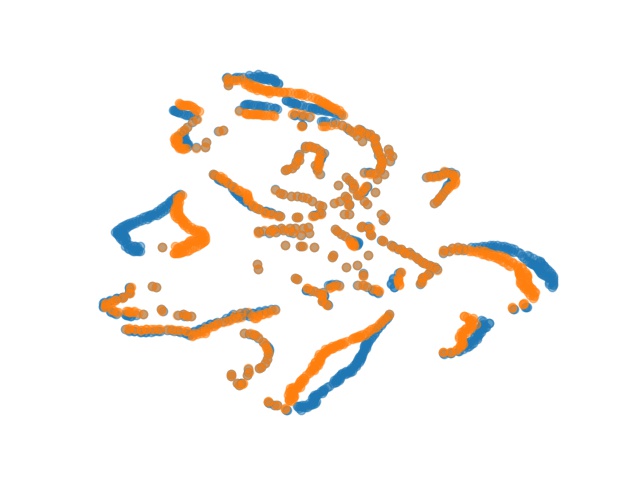} &
 \includegraphics[width=0.25\linewidth]{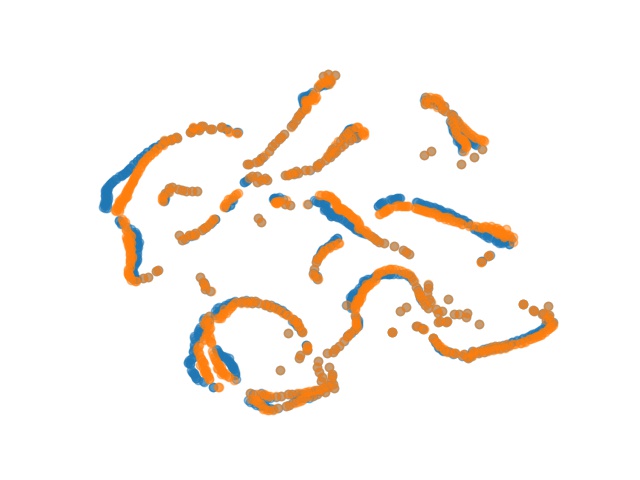} \\
 \hline
D & \includegraphics[width=0.25\linewidth]{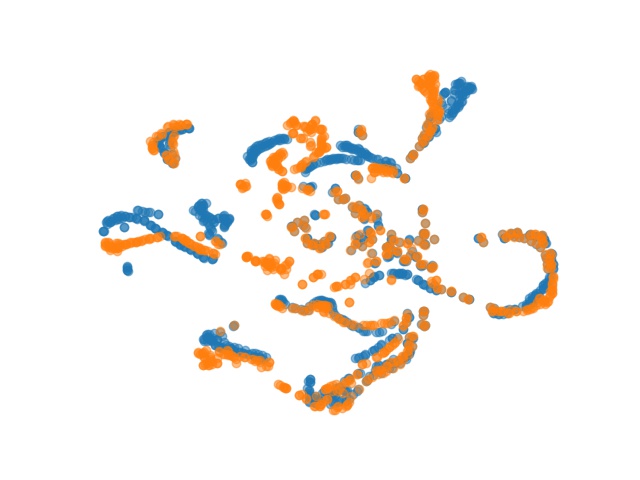} &
 \includegraphics[width=0.25\linewidth]{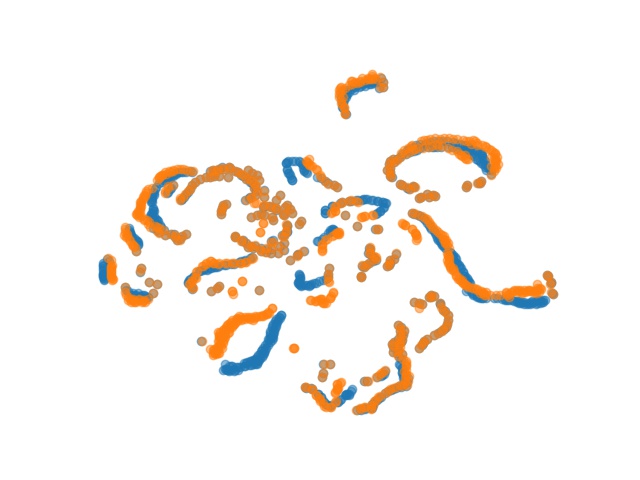} &
 \includegraphics[width=0.25\linewidth]{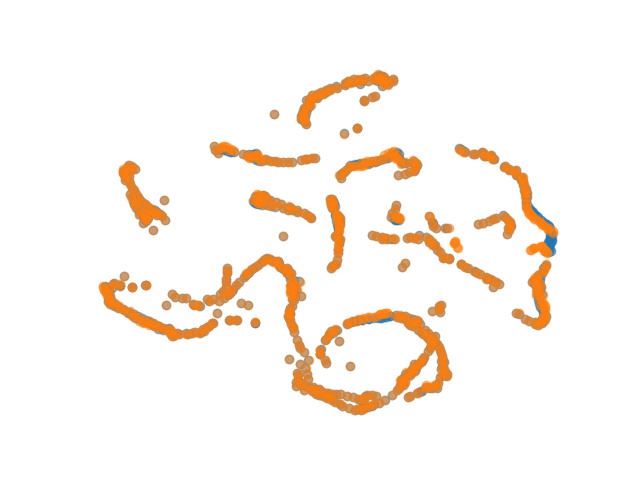} \\
 \hline
\end{tabular}
\caption{
A time-lapse t-SNE visualization of image embeddings used by our policy.
We collect images along the flying trajectory on three different tracks. 
The blue dots represent image embeddings from augmentations during training time and the orange dots represent image embeddings from test-time disturbances. A: Hue change. B: Brightness change. C: Blue boxes. D: Random Objects.
}
\label{fig: align_embed}
\end{figure}

To ensure robust feature extraction, we use contrastive learning~(Setion~\ref{sec: method-feature-learning}).
In the contrastive learning framework, the similarity loss ensures that the encoder learns the invariance between the two augmented views. As a result, the image embeddings between augmented views are aligned in the embedding space. 
We choose random convolution as the augmentation for hue changes and brightness changes;
and use random cutout-color against distractors, such as blue boxes and random objects. 
In Fig.~\ref{fig: align_embed}, we present the qualitative results of aligning image embeddings between augmentations and disturbances. 
For each of the three tracks, we collect the images along the flight trajectory of the teacher policy with either augmentations or disturbances. 
For each of the trajectories, we extract the image embeddings with the YOLO encoder and reduce the dimension to 2 with t-distributed stochastic neighbor embedding (t-SNE). 
The embeddings of each test-time disturbance are then evaluated with those of the corresponding augmentation during training. 
We can observe that the image embeddings of all the disturbances are well aligned with those of the corresponding augmentations. 
It ensures that our policy receives matching image embeddings in test time and behaves robustly. 
Thus, our policy still maintains a high success rate under all the disturbances (Table~\ref{tab:color_brightness_change}).

\subsection{Handling Noisy State}
Our policy relies on part of the drone state, including the orientation, linear velocity, and acceleration. 
The state information can be estimated using measurements from onboard sensors, such as IMUs, which are usually noisy. 
We further investigate the robustness of our policy against disturbances on the truncated sates by adding Gaussian noise $\mathcal{N}(0, std)$ individually to each component of the states, where $std$ is the standard deviation. 
Table~\ref{tab:transfer_state_disturb} shows the result. 
We can observe that the success rate decreases mildly when the standard deviation increases, which proves that our policy is robust against noises from the sensor measurements.

\begin{table}
\centering
\begin{tabular}{ccccc}
\toprule
\multicolumn{1}{c}{}                                                         & std  & Circle & Figure8 & SplitS \\ \midrule
\multicolumn{1}{c|}{\multirow{4}{*}{State Disturbance}} & \multicolumn{1}{c|}{0.04} & 1.0    & 1.0     & 1.0    \\
\multicolumn{1}{c|}{}                             & \multicolumn{1}{c|}{0.12} & 1.0    & 1.0     & 0.7    \\
\multicolumn{1}{c|}{}                             & \multicolumn{1}{c|}{0.20} & 1.0    & 1.0     & 0.4    \\
\multicolumn{1}{c|}{}                             & \multicolumn{1}{c|}{0.28} & 0.3    & 0.5     & 0.0    \\ \bottomrule
\end{tabular}
\caption{Success rates of the student policy when adding Gaussian noises to the drone states.}
\label{tab:transfer_state_disturb}
\end{table}

\section{Discussion and Conclusion}
\label{sec: conclusion}	
This work presented a method to learn deep sensorimotor policies for vision-based autonomous drone racing. 
We showed that a vision-based control policy allows predicting control commands with information extracted from images without explicitly estimating position information, trajectory planning, and tracking. 
The vision-based policy can achieve the same level of racing performance as the state-based policy while being robust against different visual disturbances and distractors. 
On the other hand, a state-based control policy is sensitive to position errors in state estimation. 
The key to achieving robust sensorimotor control is to learn well-aligned image embeddings using contrastive learning and data augmentation. 
These findings suggest that deep sensorimotor control has the potential for vision-based agile drone flight and merits further investigation. 

A major limitation of the presented work is a lack of real-world experiments to demonstrate the effectiveness and robustness of our vision-based policy. 
The deployment of the student policy on a real drone still requires further research on transfer learning or adaptive learning.
Although relaxing the need for globally-consistent position information about the drone and the gate, the student policy still relies on part of the vehicle's state to predict the control commands. We plan to tackle this in the near future by using memory-based policy representations, such as RNNs to learn hidden state representations from a history of images alone.
%
%
%
We believe our study is a stepping stone towards this goal.

\newpage

\balance

\bibliographystyle{IEEEtran}
\bibliography{references}

\end{document}